\documentclass[a4paper]{esannV2}
\pdfoutput=1
\usepackage[utf8]{inputenc}
\usepackage{amssymb,amsmath,array, microtype}

\begin{document}
\title{Challenges in anomaly and change point detection}

\author{Madalina Olteanu$^1$,  Fabrice Rossi$^1$ and Florian Yger$^2$
%
% Optional short acknowledgment: remove next line if non-needed
%\thanks{This is an optional funding source acknowledgement.}
%
% DO NOT MODIFY THE FOLLOWING '\vspace' ARGUMENT
\vspace{.3cm}\\
%
% Addresses and institutions 
1- CEREMADE, CNRS, UMR 7534, Université Paris-Dauphine,\\
PSL University, 75016 Paris, France
\vspace{.1cm}\\
2- LAMSADE, CNRS, UMR 7243, Université Paris-Dauphine,\\
PSL University, 75016
Paris, France}
%***********************************************************************
% END OF AUTHORS INFORMATION AREA
%***********************************************************************

\maketitle

\begin{abstract}
This paper presents an introduction to the state-of-the-art in anomaly and change-point detection. On the one hand, the main concepts needed to understand the vast scientific literature on those subjects are introduced. On the other, a selection of important surveys and books, as well as two selected active research topics in the field, are presented. 
\end{abstract}

% Papier de ESANN 2021 qui pourraient (eventuellement etre cités) :
% 
% Lifelong Learning from Event-based Data
% https://www2.informatik.uni-hamburg.de/wtm/publications/2021/GWLW21/ES2021-146.pdf
% mais c'est plus de l'online learning qu'autre chose ... 
%
% Enhash: A Fast Streaming Algorithm For Concept Drift Detection
%https://arxiv.org/abs/2011.03729
%
%Concept Drift Segmentation via Kolmogorov-Trees
%https://www.esann.org/sites/default/files/proceedings/2021/ES2021-93.pdf
%
%Anomalous Cluster Detection in Large Networks with Diffusion-Percolation Testing 
% https://www.esann.org/sites/default/files/proceedings/2021/ES2021-32.pdf
%
%Domain Adversarial Tangent Learning Towards Interpretable Domain Adaptation
%https://www.esann.org/sites/default/files/proceedings/2021/ES2021-103.pdf

\section{Introduction}
Real world data intensive applications are subject to the adverse effects of noise, outliers and anomalies that are common in large scale data. In addition, while shortly lived models can be built under a stationary assumption, many applications are long lived and face some form of drift in the data which can manifest as change points when the evolution is not smooth. The introduction of this paper gives a brief overview of the vast scientific literature dedicated to anomaly and change-point detection, with a focus on surveys and books. We introduce the main vocabulary and concepts needed to navigate further this literature and summarise some of the main challenges faced by practitioners. 

Anomalies and outliers are generally defined as observations that deviate in an important way from other observations. While they have been studied since the very early stages of modern statistics (see e.g. \cite{Edgeworth1887Discordant}), dealing with them remains challenging. Thousands of research papers have been published on those topics, as well as numerous surveys, review papers and textbooks. An early standard text on outlier detection from the statistical community is for instance \cite{barnett1994outliers}, originally published in 1978. Numerous problems surrounding anomaly detection were already identified (and somewhat addressed) by the first edition of this book: the unavoidable presence of outliers in large data sets, the need for multivariate outlier detection, differences between individual outliers and groups of outliers, the need to take into account the dependency structure of the data (for instance to find outliers in time series), etc. 

The need to "live with outliers" lead to the design of robust estimation methods and to the field of robust statistics (see for instance \cite{rousseeuwleroy2005robust}, originally published in 1987). The main justification for robust statistics is that non robust estimators (e.g. the mean as an estimator of the expectation) can be corrupted by a very small number of outlier observations, leading to a biased estimation bounded only by the severity of the outliers (and the sample size). This is "solved" by replacing robust variants of those estimators (e.g. the median instead of the mean). See \cite{RousseeuwMia2018Robust} for a recent survey on robust statistics applied to anomaly detection.

The robust statistics approach provides a very good illustration of the core dilemma of anomaly detection which is summarised by C. C. Aggarwal \cite{aggarwal2017outlier} with the sentence "The data model is everything". Indeed, almost all anomaly detection methods proceed by first building a model of "normal" data and then by using it to \emph{score} observations in such a way that anomalies have a high score. In statistics, models are probabilistic and the scores are inversely proportional to the likelihood of observations. More generally, scores are derived from prediction quality of the data model: a poorly predicted observation has a high score and \emph{vice versa}. Unfortunately, the presence of outliers in the data set used to build the model will generally bias it in such a way that those outliers will be considered more "normal" than they should (normal observations may also be assigned too large scores). In other words, in order to build a good anomaly detection model, one should use outlier free data!% FY : but even in this ideal case, a threshold should be chosen according to the number of outliers to expect during the test phase.

From its early statistical roots, the field of anomaly detection has evolved to encompass more \emph{ad hoc} methods from all the sub-fields of data science (e.g. deep learning, see \cite{PangEtAl2021DeepAnomalySurvey}) and to broaden it application fields, as illustrated by the data mining oriented survey from Chandola et al. \cite{ChandolaBanerjeeEtAl2009AnomalyDetection} and Aggarwal's book \cite{aggarwal2017outlier}. While numerous challenges were already identified in the statistical community, others have emerged. For instance, older statistical methods were originally focused on numerical data, while discrete data oriented methods are more recent (see \cite{TahaHadi2019CategoricalData} for a recent survey). Earlier works focusing on taking into account the dependency structure of the data were dedicated mainly to the regression case (under the Gaussian noise hypothesis) and to time series \cite{Blanquez2021TimeSeries}. This has been extended to more general structures, for instance to the classification case (especially for methods addressing \emph{label noise}, see e.g. \cite{FrenayVerleysen2014ClassificationPresence}) and to spatial structures (see for instance \cite{Shekhar2003}). Arbitrary dependencies between observations can be represented via graphs, an approach that gave birth to Graph Signal Processing (GSP) methods (see \cite{OrtegaEtAl2018GSP, DongEtal2020GSP4ML} for surveys). Graph filters can be used to remove (and therefore detect) noisy observations under arbitrary dependencies, generalising techniques developed for time series. 

More generally, recent trends include handling data with intrinsic complex structure (rather that classical data with complex dependency structure), for instance whole time series, i.e. data in which each observation is a time series, see \cite{Blanquez2021TimeSeries}. A popular case is the one of graph valued data where each observation is a graph or where the full data set is represented by a single observed graph (i.e., when the links are measured and not expert based dependency hypotheses). See \cite{Akoglu2015, MaEtAl2021GraphAnomalyDeep} for surveys. 

Change point and drift detection is a related problem that has also received a significant attention. Following the same pattern, early methods were developed by statisticians, for instance the well known CUsUM technique proposed by Page in 1954 \cite{Page1954CUsUM}. Since then, numerous papers, surveys and books have been dedicated to change point detection (see for instance \cite{basseville1993detection, tartakovsky2014sequential} for online techniques, \cite{truong2020selective} for offline ones and \cite{aminikhanghahi2017survey} for a general survey). 

The main difference between anomaly detection in a temporal context and change point detection is that the latter is generally leveraging \emph{three} data models rather than one. In essence, one compares the quality of a single model for the whole time series (or a sub-sequence) with the quality of two models, one before the tentative change point and one after. As the change point is unknown, model estimation is potentially biased: one model can be estimated using a mixture of data before the true change point and data after the true change point. This phenomenon is closely related to the one faced by anomaly detection technique, with the additional aspect that the "anomalies" are here somewhat structured (as they are produced by another model). 

In the scientific literature, the terms \emph{drift detection} are generally specific to change detection in a supervised context, especially in the classification setting. This amounts essentially to detecting when a predictive model is not adapted anymore to new data. Unsupervised methods, i.e. methods that do not assume the true labels will be known without too much delay after a prediction is made, are the closest to change point and anomaly detection ones (see \cite{GemaqueEtAl2020Drift} for a specific survey on those unsupervised methods and \cite{LuEtAl2019, AGRAHARI2021} for more general ones). 

The rest of the paper is organised as follows. We introduce the vocabulary and core concepts needed to understand the challenges of anomaly and change point detection in Section \ref{section:core:concepts}. Section \ref{section:data:model} discusses C. C. Aggarwal's summary of the main difficulty faced by all techniques: "The data model is everything". Section \ref{section:challenge} concludes the paper by presenting two of the main current challenges in the field. 

\section{Core concepts}\label{section:core:concepts}
\subsection{Anomaly, noise and outlier}
Anomalies and outliers are generally defined in plain English using arguably vague sentences such as
\begin{itemize}
    \item an anomaly "\emph{appears to deviate markedly from other members of the sample in which it occurs}" \cite{Grubbs1969};
    \item an outlier is "\emph{an observation which deviates so much from the other observations as to arouse suspicions that it was generated by a different mechanism}"~\cite{hawkins1980identification};
    \item an outlier is "\emph{an observation (or subset of observations) which appears to be inconsistent with the remainder of that set of data}" \cite{barnett1994outliers}.
\end{itemize}
Those informal definitions should be considered as guiding principle to define mathematically sound and operational definitions, as we will see below. In particular, Chandola et al. write in \cite{ChandolaBanerjeeEtAl2009AnomalyDetection} "\emph{Anomalies are patterns in data that do not conform to a well defined notion of normal behavior.}" which constitutes already a step towards more formal definitions. 

Notice that we do not distinguish between anomaly and outlier, but there is some value in being more precise in some contexts. Aggarwal proposes in \cite{aggarwal2017outlier} to distinguish noise from anomaly by the intensity of the deviation from normality: noisy data are more normal than anomalies. Then he proposes to call outliers noisy data as well as anomalies (possibly distinguishing weak outliers from strong outliers). 

This distinction can be illustrated in probabilistic terms with the classical Gaussian distribution. Let us assume that a quantity is measured with an additive measurement noise distributed according to a standard normal distribution and that the true quantity should be 100 (in some adapted measurement unit). Then any measurement between 98 and 102 will be considered normal because roughly 95\% of measurements made on a true quantity of 100 will fall in this interval. A measurement of 103 could be considered as noisy, as a true quantity of 100 shall lead to a measurement larger or equal to 103 only in 0.13\% of the observations. Then a measurement of 105 could be considered as anomalous because observing values larger or equal to 105 should happen only once for more than 3 millions measurements on average. 

This basic example illustrates the main difficulties of anomaly detection: we need to chose a data model (here the standard normal distribution and the fact that the true quantity is 100) as well as decision thresholds. 

\subsection{Type of anomalies}
Chandola et al. \cite{ChandolaBanerjeeEtAl2009AnomalyDetection} sort anomalies into three categories or types. The simplest case is the one of \emph{point anomalies} where a single observation can be classified in isolation as an anomaly. This type of anomaly is the main focus of most of the methods. In statistical and machine learning terms, it is associated to the classical independence hypothesis between observations. 

When the observations are statistically dependent, the notion of anomaly should be revised. Indeed, the expected value of an observation is in this case dependent from the values of other observations. Then the status of an observation (normal or anomalous) cannot be decided in isolation. Such anomalies are called \emph{contextual anomalies} or \emph{conditional anomalies}. 

The most well known case of contextual anomalies is the one of time series \cite{Blanquez2021TimeSeries} but numerous extensions have been considered, some recently. It should be first noted that the terms \emph{time series} generally refer to temporal data with numerical values, while there is also an important literature on discrete sequences, i.e. time series with categorical values (see \cite{ChandolaEtAl2012DiscreteSequences} for a survey). Spatial data have also been studied, for instance in \cite{Shekhar2003}. 

As pointed out in the introduction, graphs can be used to represented fully arbitrary dependencies between observations and denoising filters from signal processing can be generalised from the time domain to the graphical one, leading to the field of graph signal processing (GSP) \cite{OrtegaEtAl2018GSP, DongEtal2020GSP4ML}. When the graph is observed, one can use it as a dependency hypothesis, leveraging GSP or other approach, or, on the contrary, one can question it, considering edges as potential outliers. This leads to a generalised notion of contextual anomaly specific to graph anomaly detection \cite{Akoglu2015, MaEtAl2021GraphAnomalyDeep}. 

The third category of anomalies is the one that concerns groups of observations. In point or contextual anomalies, a \emph{single} observation is considered anomalous when confronted to a subset of related observations (which are themselves normal). In \emph{collective} anomalies, a subset of observations is considered \emph{as a whole} to be anomalous even if each observation is normal considered in isolation (or within its natural context). A typical example is a faulty sensor that stops updating for a short time period and then resumes normal operation: the fact that the sensor reports several identical values while possible might be a hint that something is not working as expected, even if the value is perfectly valid. In addition to the temporal context, collective anomalies are frequently searched for in spatial data and in graphs (see e.g. \cite{LarrocheEtAl2020}). 

In practical applications, such as computer intrusion detection (see e.g. \cite{AHMED201619}), it is common to look for several types of anomalies, depending on the hypotheses on data generation. In addition, data tend to be collected more and more thoroughly leading to complex data sets such as temporal network data, which combine naturally a complex structure and non trivial dependencies. 

\subsection{Change and drift detection}
Change-point detection relates to anomaly detection, in particular to contextual and collective anomalies. The main conceptual difference between both is that data are "normal" on both sides of a change point, and are only considered anomalous \emph{from the point of view of the other model}.  Change-point analysis implies the data to be available in a sequential manner, and is therefore more common for temporal or streaming data, or more generally for ordered data (see the following section). 

Generally designed in an unsupervised context, change-point detection builds upon the idea that the underlying distribution of the data is abruptly changing at some unknown instants. While the first historical methods focused on detecting changes in the mean value (as in CUsUM \cite{Page1954CUsUM} discussed in the introduction), subsequent developments proposed numerous modelling frameworks, both parametric and nonparametric. 

In the supervised learning context, an abrupt  change in the data distribution is generally called a \emph{concept shift}. Because of the asymmetric nature of the data, with "inputs" $X_t$ and "outputs" (a.k.a. predictions) $Y_t$, research on concept shift distinguishes different types of change, namely in marginal distributions ($\mathbb{P}(X)$ or $\mathbb{P}(Y)$) or in conditional distributions ($\mathbb{P}(X|Y)$ and $\mathbb{P}(Y|X)$). Covariate shift~\cite{sugiyama2012machine} and prior probability shift are the terms used to refer to changes in the marginal of $X$ or $Y$ respectively. The main focus of research is generally \emph{true concept shift}, i.e. changes over time in $\mathbb{P}(Y|X)$, including the difficult case of handling new values for $Y$ in the classification setting. 
The term data set shift~\cite{quinonero2008dataset} is also used in the literature and can encompass many diverse situations ranging from the aforementioned concept drift to sample selection and domain shift.

While there is not a perfect consensus on the vocabulary, it is generally admitted to use \emph{shift} for abrupt changes and \emph{drift} for smoother change in the data distribution (see e.g. the discussion in \cite{AGRAHARI2021}). Notice however, that in the data mining community, especially in data stream literature, the dominant term is \emph{concept drift}, in a way that encompass both abrupt and gradual changes, and that even extends to the unsupervised context (see \cite{GemaqueEtAl2020Drift}). 

%% discuss multiple change detection?: not enough space :-)

\subsection{Operational hypotheses}
The conditions under which anomaly or change detection is performed can have a significant impact on the design and the performances of the detection methods. 

While very uncommon, the case of supervised learning should be mentioned for anomaly detection. Indeed in some application contexts such as fraud or computer intrusion detection, it may be possible to collect a data set with labelled examples combining (a lot of) normal examples and (a small set of) anomalous examples. In this case, the problem is a standard but difficult supervised learning one. The difficulties come from the  unbalanced nature of the data (collecting examples of anomalous behaviour is generally difficult) and from the ill-posed nature of the classification: while the normal data class is well defined, the anomalous data form a collection of unrelated examples that can exhibit vastly different characteristics (in contrast to change point or drift detection settings where both data distribution are supposed to be somewhat consistent). In addition, the normal data set can be contaminated by undetected anomalies. Variations over the fully supervised learning context include the also classical semi-supervised one, with a specific variant where only one class (the anomalous one) is labelled. We refer the reader to Chapter~7 of \cite{aggarwal2017outlier} for further details. 

In practice, anomaly and change point detection will therefore be mostly conducted under the unsupervised learning paradigm (to the point where most surveys discuss only unsupervised techniques). On the contrary, concept drift detection will be mostly conducted under the supervised learning paradigm. 

Another important aspect is specific to temporal (or ordered) data and applies to both change and anomaly detection. Models can be applied in an on-line way or in an off-line way. In the former, the data appear as a stream and the goal is to take a decision on the last observation using only the past observations. This can be slightly relaxed by allowing some delay between the observation and the decision, which enables the method to see some observations that occurred after the one under analysis. In the off-line mode, the data is fully observed before applying the method. A classical trade-off between the modes are accuracy versus delay: on-line methods are generally less accurate than off-line ones, but the latter cannot be used in a streaming context or to react quickly to new observations. 

%% Do we include score versus class here or in the next section? Let's keep the current organisation
% FY : the issue with modelling the outliers as a class being addressed just before, it may be enough (?)

\section{"The data model is everything"}\label{section:data:model}
\subsection{A unifying view}
As pointed out in \cite{aggarwal2017outlier} one can summarise the whole field of anomaly detection as follows: "\emph{Virtually all outlier detection algorithms create a model of the normal patterns in the data,
and then compute an outlier score of a given data point on the basis of the deviations
from these patterns.}" As discussed above, this generalises to some extent to change point detection and to concept drift detection. In those cases, we have a "current" data model which is confronted either to some new observations or directly to another data model. In these settings, models are compared via observations, by assessing which model is better at describing one or several observations. Thus anomaly and change detection methods differ by the assumptions they make on the data as translated into the model building strategy, and by the aggregation level at which they compare entities of interest (a single observation versus a model, a collection of observations versus a model, two models). 

\subsection{Consequences}
As already discussed in the Introduction, this "model and score" view emphasises the core difficulty of anomaly and change detection: in general, the data model will be adjusted to the observations \emph{blindly}, that is without knowing in advance whether a given observation is normal or not. Thus model fitting must be somehow "robust" to the presence of outliers (not necessarily in the robust statistics sense). This is surprisingly difficult as even "simple" tasks such as computing a mean or a covariance matrix are biased in presence of anomalies~\cite{rousseeuwleroy2005robust}. 

In addition, owing to the unsupervised nature of the problem (in most of the cases), model choice is difficult. In essence, we are in a typical case of a ill-posed model. If we come back to the very basic example of a Gaussian noise described above, results would be quite different using another noise distribution such as a Laplace  one. For instance using a Laplace distribution with unitary variance, centred on 100, the probability of observing 105 or more is larger than 1/2500, compared to $1/(3.5\times 10^6)$ for the standard normal distribution. In a sense the Laplace noise model is more tolerant to extreme values than the Gaussian one. More generally, the nature of some observations is very likely to depend on expert hypotheses on the data generation process and it seems a bit naive to hope for fully automated generic models. 

Moreover, the vast majority of the methods output an anomaly score or a comparable quantity in change detection approaches (e.g. model dissimilarities). In a decision oriented setting, this continuous quantity must be turned into a binary one: is this observation an anomaly? is there a shift in the data distribution? While researchers tend to focus on score evaluation, using for instance the area under the ROC curve (AUC) as an integrated quality metric, practitioners generally need decisions. It is common to read that ranking the observations based on their score allows one to avoid setting a decision threshold. However ranking is putting the weight of the decision on the analyst shoulders, as he/she will have to decide where to stop in the list of ranked observations. Even worse, the stopping decision could be driven in this case by operational considerations (such as human resources  available to investigate the anomalies). While those are valid considerations, they should be explicitly stated. In summary, thresholding scores into decisions is part of the model fitting process and should not be ignored. Notice that the discussion applies almost as is to change point detection: fixing the decision threshold is equivalent to fixing a number of anomalies (on a static data set), whose counterpart in change detection is the number of change points. 

\section{Challenges and opportunities}\label{section:challenge}
As summarised above, anomaly detection and change detection are somewhat ill-posed problems (as e.g. clustering \cite{pmlr-v27-luxburg12a}) and expert input and monitoring is necessary to obtain meaningful results. Evaluating objectively the proposed methods remains also a difficult problem for the same reasons. 

\subsection{Interpretability and visualisation}
On the best way to engage users in machine learning algorithm monitoring and steering is via information visualisation techniques \cite{EndertEtal2017}, especially when the goal is explicitly to enhance the trust of the user into the results provided by the algorithm \cite{ChatzimparmpasEtAl2020}. As far as we known there is unfortunately no general survey on visualisation techniques directly targeting anomaly or change detection, but two specific application contexts have been reviewed: network monitoring \cite{Zhang2017} and user behaviour \cite{ShiEtAl2022}. 

More generally, there is a strong need for interpretable detection. A recent survey \cite{Panjei2022} presents the state-of-the-art in this direction. It outlines several strategies to provide anomaly explanations, mainly score unification (that brings comparability between detection methods), outlying attribute identification and causal interaction among anomalies ($x$ is an anomaly because $y$ appeared before in the data set). 

Both visualisation and interpretability remain relatively new in the literature but are also considered as very important for the applicability of detection methods in real world application, as argued by e.g. this position paper \cite{Riveiro2016Importance} and this survey \cite{MaEtAl2021GraphAnomalyDeep}. These topics are actively researched as exemplified by Hinder et al. paper \cite{ESANN22-contrasting} is the present volume. 

\subsection{Benchmarking}
Unsupervised models are notoriously difficult to evaluate \cite{pmlr-v27-luxburg12a} as a consequence of their ill-posed nature: each model uses generally different hypotheses about the data generation process, with an associated quality metric for which it is optimal (but of course not for other metrics). The gold standard is to refer to a collection of \emph{labelled} data sets where the ground truth is known with reasonable certainty (and representative of the task at hand). This is surprisingly difficult to achieve in anomaly and change detection settings, generally for cost reasons as, by definition, expert based labelling is needed to build the ground truth. This is particularly costly because the anomalies are by essence rare: in order to label enough anomalies, very large data sets are needed. An example of evaluation under this gold standard is included in Coussirou et aL. paper \cite{ESANN22-oilsystem} in this volume. In some situations, it is possible to generate anomalies or changes using artificial models or human intervention. This is illustrated by Bel-Hadj et al. paper \cite{ESANN22-railwaybridge} in the present volume. But this is generally not the case, as discussed for instance by \cite{AHMED201619} in the case of classical network intrusion benchmarks.

\begin{footnotesize}

\bibliographystyle{abbrv}
\bibliography{biblio}

\end{footnotesize}

\end{document}